\RequirePackage{fix-cm}
\documentclass{svjour3}                     
\smartqed  
\usepackage{graphicx}
\usepackage{todonotes}
\usepackage{url}
\usepackage{xcolor}

\begin{document}

\title{Visual link retrieval and knowledge discovery in painting datasets
}

\titlerunning{Visual link retrieval and knowledge discovery in painting datasets
}        

\author{Giovanna Castellano \and Eufemia Lella \and Gennaro Vessio 
}


\institute{G.C. and G.V. \at
            Department of Computer Science,  University of Bari ``Aldo Moro'', Bari,  Italy\\
            \email{gennaro.vessio@uniba.it}           
           \and
           E.L. \at
            Innovation Lab, Exprivia S.p.A., Molfetta, Italy\\
}

\date{Received: date / Accepted: date}

\maketitle

\begin{abstract}
Visual arts are of inestimable importance for the cultural, historic and economic growth of our society. One of the building blocks of most analysis in visual arts is to find similarity relationships among paintings of different artists and painting schools. To help art historians better understand visual arts, this paper presents a framework for \textit{visual link retrieval} and \textit{knowledge discovery} in digital painting datasets. Visual link retrieval is accomplished by using a deep convolutional neural network to perform feature extraction and a fully unsupervised nearest neighbor mechanism to retrieve links among digitized paintings. \textit{Historical} knowledge discovery is achieved by performing a graph analysis that makes it possible to study influences among artists. An experimental evaluation on a database collecting paintings by very popular artists shows the effectiveness of the method. The unsupervised strategy makes the method interesting especially in cases where metadata are scarce, unavailable or difficult to collect.

\end{abstract}

\section{Introduction}
\label{intro}

Visual arts play a strategic role for the cultural, historic and economic growth of our society \cite{leavy2017handbook}. They stimulate interest and can change the way we look at the world around us. They tell stories that words cannot capture. Visual arts are also vital for children's learning, as they can help students form their creativity while developing their personality \cite{van2015enhancing}. 

In the last years, due to technological improvements and the drastic decline in costs, a large scale digitization effort has been made, leading to a growing availability of large digitized fine art collections \cite{windhager2018visualization}, for example WikiArt\footnote{\url{https://www.wikiart.org}}~and the MET collection.\footnote{\url{https://www.metmuseum.org/art/collection}} This availability, coupled with the recent advances in Computer Vision and Pattern Recognition, has opened new opportunities for computer science researchers to assist the art community with automatic tools that analyze and help further understand visual arts. A deeper understanding of visual arts has the potential to make them more accessible to a wider population, both in terms of fruition and creation, and to enrich human-computer interaction, which is often inspired by artistic paradigms. 


Understanding high-level semantic attributes of a painting, such as content and meaning, inherently falls within the domain of human perception. In fact, it originates from the ability to recognize meaningful low-level patterns, such as the composition of shapes, colour and texture features, which are visually perceived by the human eye. Computer Vision techniques, in particular Convolutional Neural Networks (CNNs) \cite{liu2017survey}, are very effective to tackle the problem of learning useful high-level representations from the low-level colour and texture features. These representations can assist in various visual art related tasks, ranging from object detection in paintings \cite{crowley2014search} to artistic style categorization \cite{van2015toward}.

\begin{figure}
    \centering
    \includegraphics[width=.3\columnwidth]{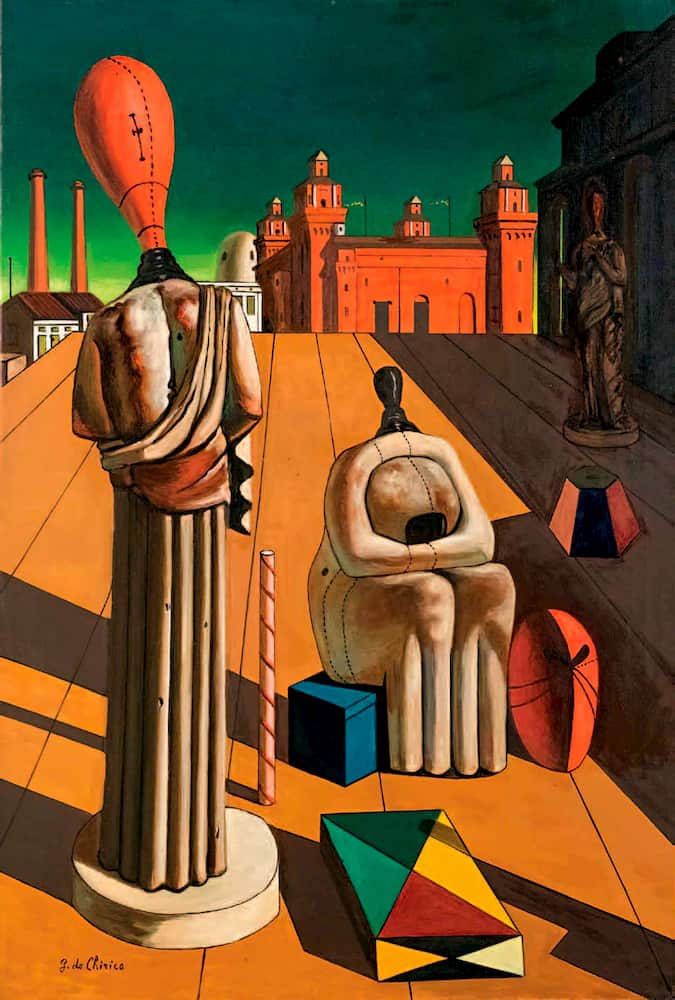}
    \caption{``The Disquieting Muses'' by Giorgio De Chirico (1917). This metaphysical painting is clearly inspired by classical Greek culture and, albeit in a provocative way, displays influences from African traditions that inspired several painters, including Pablo Picasso, in the same period.}
    \label{fig:muses}
\end{figure}

One of the building blocks of most analysis in visual arts is to find similarity relationships, i.e.~link retrieval, among paintings of different artists and painting schools. These relationships can help art historians discover and better understand influences and changes from an artistic movement to another. Art experts, in fact, rarely analyze artworks as isolated creations, but typically study paintings within broad contexts, involving influences and connections among different schools (see Fig. \ref{fig:muses}). Traditionally, this kind of analysis is done manually by inspecting large collections of human annotated photos. However, manually searching over thousands of pictures, spanned across different epochs and painting schools, is a very time consuming and expensive process. 

Along this direction, the contribution of this paper is two-fold:
\begin{enumerate}
    \item A deep learning-based framework is proposed for \textit{visual link retrieval} within large digitized collections of paintings, based only on simple image queries. Although the proposed approach is based on previous well-known methods, as far as we know it represents the first application of an unsupervised deep learning-oriented approach to this kind of problem;
    \item The proposed method not only provides those images that are more similarly linked to the input query, but also allows the user to study historical patterns by means of graph analysis. In fact, by applying graph measures on the network built upon the links obtained, the proposed method performs a form of \textit{historical knowledge discovery}.
\end{enumerate}



The rest of this paper is structured as follows. Section 2 discusses related work. Section 3 presents the proposed method. Section 4 reports the results of an experimental evaluation on a dataset of very popular painters. Section 5 concludes the paper and sketches future developments of the present research.

\section{Related work}
\label{related}

Classification and retrieval of artistic images are very common tasks in the field of automatic art analysis. In the past years, such tasks have typically been addressed using hand-crafted features \cite{carneiro2012artistic,khan2014painting,shamir2010impressionism}. However, despite the promising results of feature engineering techniques, early attempts were affected by the difficulty of capturing explicit knowledge about the attributes to be associated with a particular artist or artwork. The difficulty arises because this knowledge is associated with implicit and subjective expertise human experts may find hard to verbalize. 

Conversely, several successful applications in a number of Computer Vision tasks (e.g.,  \cite{acharya2019deep,castellano2020crowd,howard2017mobilenets}) have shown that representation learning is an effective alternative to feature engineering, especially if combined with deep neural network architectures. One of the main reasons of the recent success of deep CNNs in solving tasks too difficult for classic algorithms is the availability of large human annotated datasets, such as ImageNet \cite{russakovsky2015imagenet}. A model built on a large dataset often provides sufficiently general visual features that can be profitably used, through \textit{transfer learning}, for more specific tasks. In particular, deep neural networks have been successfully applied to solve various tasks related to visual arts.

Several works have focused on object recognition and detection in artworks \cite{cai2015cross,cai2015beyond,crowley2016art,gonthier2018weakly,tan2016ceci,wilber2017bam}. A first attempt to use deep neural networks for object recognition in visual arts was presented in \cite{crowley2014search}. In the work, Crowley and Zisserman developed a CNN-based system that can learn object classifiers from Google images and use them to find previously unseen objects in a large database of paintings. 

Another task frequently addressed by computer science researchers in the domain of visual arts is learning to recognize artists by their style. One of the first works in this context is the research presented in \cite{van2015toward}: van Noord et al.~proposed PigeoNET, a CNN trained on a large collection of paintings to perform the task of automatic artist association based on visual characteristics. These characteristics can also be used to reveal the artist of a precise area of an artwork, in the case of multiple authorship of the same work. Encouraging results from the application of deep CNNs to artistic style classification have been recently reported in \cite{cetinic2018fine,mao2017deepart,sandoval2019two}. Other works, such as \cite{strezoski2017omniart}, have also experimented with CNN models trained with additional data, particularly time period, reporting better results. 

The notable contributions previously described confirm the applicability of a deep learning-based strategy to another task related to the visual art domain, i.e.~the problem of visual link retrieval in painting datasets. Indeed, this task has not been extensively investigated before. Recently, deep learning-based approaches have been proposed to retrieve common visual patterns shared among paintings. In \cite{seguin2016visual}, Seguin et al.~compared a classic bag-of-words method and a pre-trained CNN in predicting pairs of paintings that an expert considered to be visually related to each other. The authors have shown that the CNN-based method is able to surpass the more classic one. The authors used a supervised approach in which the labels to be predicted were provided manually by human experts. This manual  annotation of images is a slow, error-prone and highly subjective process. Conversely, a fully unsupervised learning approach would avoid this cumbersome process. In \cite{shen2019discovering}, Shen et al.~used a deep neural network model to identify near-duplicate patterns in a dataset of artworks attributed to Jan Brueghel. The key technical insight of the method is to adapt a deep standard feature to this task, perfecting it on the specific art collection using self-supervised learning. Spatial consistency between adjacent feature matches is used as a supervisory fine-tuning signal. The fitted function leads to a more accurate style invariant match and can be used with a standard discovery approach, based on geometric verification, to identify duplicate patterns in the dataset. The method is self-supervised, which means that the training labels are derived from the input data. Conversely, in our method we rely on a completely unsupervised approach, where there is no intrinsic class labeling and the method should find the pattern on its own.

An unsupervised approach to finding similarities among paintings was proposed by Saleh et al.~\cite{saleh2016toward}, based on traditional hand-crafted features. They trained discriminative and generative models for the supervised task of classifying painting style to ascertain what kind of features would be most useful in the artistic domain. Then, once they found the most appropriate features, i.e.~those that achieve the highest accuracy, they used these features to judge the similarity between paintings using distance measures. This work suggested that high-level semantic features, such as those that can be extracted with a deep neural network, may pave the way for capturing the subjective perception with which a human being judges complex visual concepts.
 
In line with these ideas, this paper proposes a method for visual link retrieval that works in a completely unsupervised way, without the need for human annotations to painting images. The method relies solely on visual attributes that are automatically learned by a deep neural network from the painting collection itself. In this way, a computer-automated suggestion of influences between artists is obtained. The ability to rely only on visual patterns, without any human intervention, makes the proposed approach particularly desirable especially when it is difficult to collect textual metadata, which may be scarce or unavailable. A preliminary sketch of this approach has been described in \cite{castellano2020towards}, where its effectiveness was first investigated. The present work significantly extends that previous work, reporting the results of an empirical evaluation carried out with art experts. Moreover, this work introduces an additional graph-based analysis of the painters' network obtained from the visual links retrieved. The analysis of the network structure provides an interesting insight into the influences among artists that can be considered the result of a novel knowledge discovery process.
  
It is worth noting that, as done in this paper, some latest image retrieval methods~\cite{kalaiarasi2013visual,minu2014semantic,nagarajan2012machine}, do not rely on human annotations. The main assumption of the present work is that identifying similarities in digital artworks and digital images using CNNs are similar tasks. In other words, both digital artworks and digital images will be treated in the same way by a CNN, when this is used for feature extraction. In fact, similarities or dissimilarities among artworks are assumed to be transferred to the corresponding images when they are digitized.

\section{Proposed method}
\label{proposed}

The proposed method assumes the availability of a large collection of digitized paintings of different artists and genres, such as those collected today in several online museum and art gallery databases. The goal is to transform the raw pixel images into a new, numerical feature space in which to look for similarities among paintings. These similarities can be used to provide semantic links among paintings and to build a network of influences among painters. 

The algorithm involves the following steps:
\begin{enumerate}
    \item Given a digital painting image dataset, a pre-processing step is performed to resize and normalize the images;
    \item A pre-trained CNN is used to automatically extract higher-level features from the pre-processed digital painting images;
    \item The resulting high dimensional representation is embedded into a more compact feature space by applying Principal Component Analysis (PCA);
    \item Similarities among paintings, i.e.~visual  links, are obtained through a distance computation in a completely unsupervised Nearest Neighbor (NN) fashion;
    \item Given a query image, its $k$ nearest neighbors are retrieved by the system in the embedded feature space;
    \item Once the nearest neighbors for all painters have been collected, an undirected graph is constructed to express the connections between artists; 
    \item Finally, some graph measures are applied to this graph to describe its topological properties (which can be reformulated as artistic influences).
\end{enumerate}
These steps are described in more detail in the next subsections. 
A list of acronyms and symbols used to describe the method is provided in Table \ref{tab:symbols}.

\begin{table}[t]
\caption{List of acronyms and symbols.}
\label{tab:symbols}       
\begin{tabular}{ll}
\hline\noalign{\smallskip}
Abbreviation & Description \\
\noalign{\smallskip}\hline\noalign{\smallskip}
$B(v)$ & Betweenness centrality of node $v$ \\
$C(v)$ & Closeness centrality of node $v$ \\
CNN & Convolutional Neural Network \\
$deg(V)$ & Degree of node $v$ \\
$G(V, E)$ & Graph $G$ of $V$ nodes connected by $E$ edges \\
$k$ & Number of nearest neighbors \\ 
$\ell_2$ & Standard distance between points in space \\
NN & Nearest Neighbors algorithm \\
PCA & Principal Component Analysis \\
$\texttt{p}, \texttt{q}$ & Painting images \\
$s, t, u, v$ & Graph nodes (i.e., artists) \\
VGG16 & Well-known pre-trained CNN \\
\noalign{\smallskip}\hline
\end{tabular}
\end{table}

\subsection{Visual link retrieval}
\label{visual}

\begin{figure}
    \centering
    \includegraphics[width=.85\columnwidth]{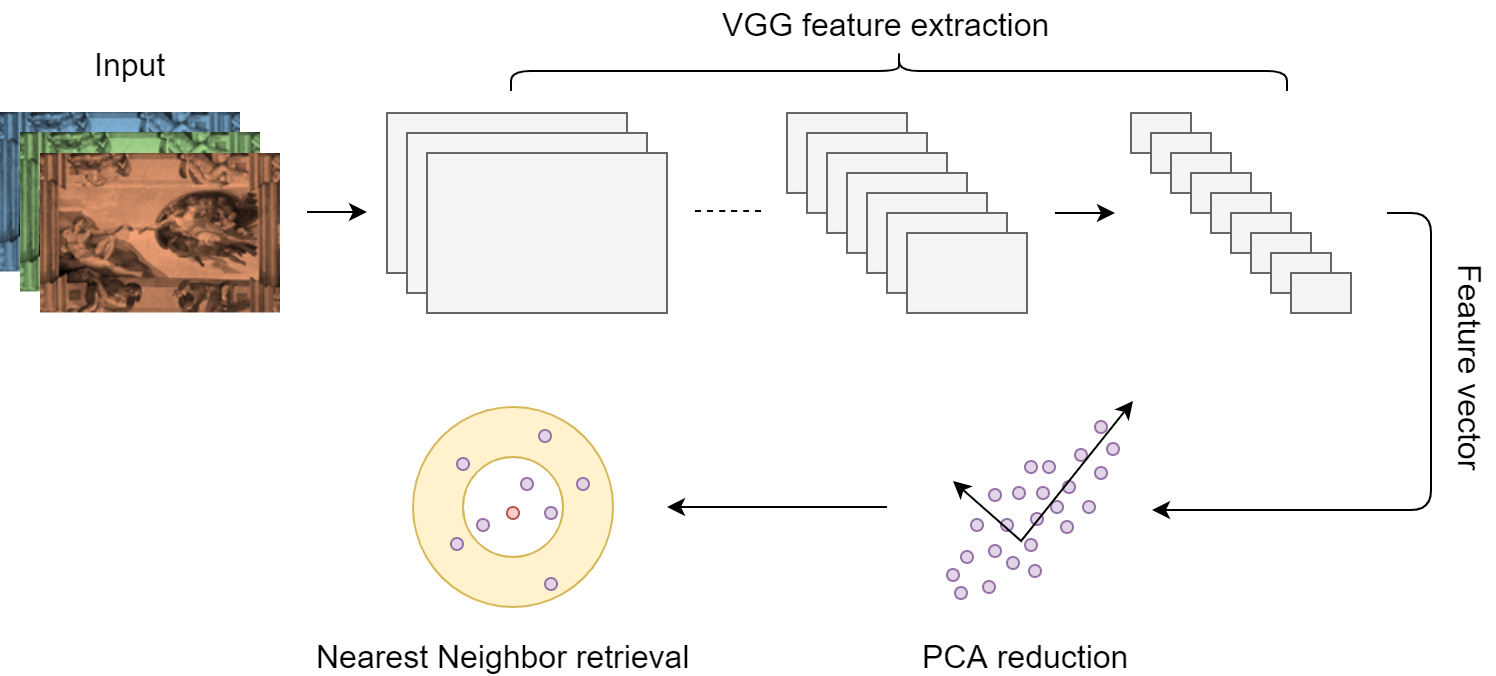}
    \caption{Workflow of the proposed retrieval method. The pre-processed images are fed into VGG16, which is used to perform feature extraction. The dimensionality of the resulting feature space is reduced by applying PCA. Visual links are retrieved in the reduced feature space using the unsupervised Nearest Neighbor.}
    \label{fig:workflow}
\end{figure}

A general scheme of the proposed framework for visual link retrieval is shown in Fig.~\ref{fig:workflow}. In order to obtain meaningful representations of visual attributes of paintings, \textit{transfer learning} is used based on a state-of-the-art deep CNN, i.e.~VGG16 \cite{simonyan2014very}, pre-trained on the very large ImageNet dataset \cite{russakovsky2015imagenet}. The input to the system is represented by $224 \times 224$ three-channel painting images, normalized in the range $\left[ 0, 1 \right]$: this is the typical input expected by VGG16. The main assumption is that if the original dataset is large and general enough, then the weights learned by the network on this set of data can be used for new, even completely different image datasets. Note that, although ImageNet does not share the same type of (painting) images used in the present work domain, transfer learning has been preferred over training the network from scratch, as the proposed method assumes no \textit{a priori} knowledge of the specific domain. Conversely, as stated earlier, the method assumes that digital images and digital arts will be treated in the same way by the CNN.

VGG16 is a well-known CNN architecture that adopts $3 \times 3$ convolution and $2 \times 2$ max pooling throughout the network. It follows a classic scheme in which pairs of convolutional layers are followed by a max pooling layer, and so on, for a total of 16 weight layers. All hidden layers are equipped with the ReLU activation function. VGG16 was preferred to other more sophisticated deep networks as it presents a classic scheme, common to more modern CNNs, thus allowing to demonstrate the applicability of the method using a simple and general scheme.

In the proposed method, the deep network is used to extract meaningful features from the pixel values of low-level raw images. Transfer learning is achieved by using the common practice of ignoring the fully-connected layers stacked on top of the convolutional base and extracting the output features from the last max pooling layer. The network is able to construct a hierarchy of visual features, starting with simple edges and shapes in the earlier layers up to higher-level concepts such as objects and complex shapes in the following layers. This approach is therefore suitable for obtaining high-level, semantic representations for the problem at hand.

Once the features extracted by the deep network are flattened, they have a very high dimensionality (i.e., $25,088$) which prevents the use of distance measures. For this reason, this high dimensional feature space is first transformed into a more compact low dimensional representation by adopting PCA \cite{jolliffe2016principal}. Several works (e.g., \cite{athiwaratkun2015feature,ren2012unsupervised}) make direct use of the last feature map of a CNN; on the contrary, PCA adds an extra component to the model, thus requiring extra time. However, it allows for a more accurate and efficient search of visual links with the following Nearest Neighbor searching mechanism. One one hand, NN is well-known for suffering from the ``course of dimensionality'' which makes distance measures meaningless as the dimension of data increases. In fact, when dimensionality increases, the volume of the space increases so much that the available data become sparser and sparses, ending with most of the points lying at the boundary of the (hyper) cube. On the other hand, the neighbor search of the algorithm grows linearly with the dimension (and the number) of data points. When data are fixed, PCA only needs to be performed once; instead, the NN prediction must be run for each query. In light of this, the computational cost of PCA becomes marginal and unimportant. More precisely, the training time can be technically ignored, as it is only conducted once offline. Regarding the prediction time, it depends on the implementation of NN. The more na\"{i}ve implementation of the neighbor search involves calculating the brute force of the distances between all pairs of points in the dataset. For $N$ samples in $D$ dimensions, this approach scales as $O(DN^2)$. Assuming that the number of neighbors that the algorithm has to retrieve, $k$, is fixed, the time complexity is $O(NDk)$. To address the computational inefficiencies of the brute force approach, a variety of tree-based data structures have been proposed and can be used.

To achieve a good compromise between representation power and dimensionality, the original feature space is projected onto a reduced space of $50$ features (referred to as \textit{principal components}). The total variance within the new feature space is not uniformly distributed among features, but redistributed in an unequal way: the first principal components explain most of the variance of the new feature set. In this way, it is possible to drastically reduce the dimensions of the original feature space, without sacrificing too much information.

The final search for visual links among paintings is performed in the reduced feature space in a fully unsupervised  fashion using a Nearest Neighbor mechanism. In other words, for each query point $\texttt{q}$ the methods returns the $k$ data points \textit{closest} to $\texttt{q}$. ``Closeness'' implies a metric which, as in PCA, corresponds to the usual $\ell_2$ distance \cite{grobe1994student}:
\[
\ell_2(\texttt{q}, \texttt{p}) = \sqrt{\sum_{i=1}^{D} \left( q_i - p_i \right)^2 },
\]
where $\texttt{q}$ is the query point, $\texttt{p}$ is any other data point, and $D$ is their dimensionality. In this way, for each query, the $k$ most similar paintings are provided by the system. It is worth noting that, when using a search query from a particular artist, other paintings from the same artist are excluded from the search, otherwise obvious, self links are likely to be retrieved. 
This is different from retrieving near-duplicate images (see, for example, \cite{thyagharajan2020review} and \cite{thyagharajan2018pulse}), since if the same artist did a different painting, this will not be retrieved as similar or near-similar work. The concept of near-duplicate image retrieval does not apply to the same artist in this paper, as the secondary goal of the proposed method is to find similarities among different artists, thus retrieving influences among them.

As stated earlier, relying on a completely unsupervised approach makes the proposed method simple and practical, as it excludes the need to acquire visual link labels, which can be very difficult to collect.

\subsection{Knowledge discovery}
\label{discovery}

Once the nearest neighbors of all painters' artworks are collected, for each painter $v$ the method identifies the most recurring linked painter $u$ as the one whose works were found most visually related to the works of $v$. Using this information, it is possible to build an undirected graph $G (V, E)$, whose nodes are the artists taken into account, while edges express the similarity between their paintings. More precisely, for each $v \in V$, being $V$ the set of painters, the artworks made by $V$, $\texttt{q}_1, \texttt{q}_2, \ldots$, are taken into consideration. Then, for each artwork $\texttt{q}_i$, the top one visually linked painting $\texttt{p}_i$, belonging to another artist $u \in V$, $u \neq v$, is considered. Iterating over all $\texttt{q}_i$ allows the system to provide a ranked list of artists, the works of which have been retrieved as being the most visually linked with the works of $v$. This list is ranked since these artists may be present at different frequencies. Finally, an edge $e \in E$ is assumed to exist between $v$ and $u$, provided that $u$ is the most recurring artist in the previously mentioned list of artists linked to $v$. The edge is undirected because if $v$ is similar to $u$, then $u$ is similar to $v$ (so, no time relationship is explicitly taken into account). Note that the graph $G$ does not reduce to a collection of disconnected node pairs. In fact, the existence of an edge between $v$ and $u$ does not prevent $u$ from being more visually linked to another artist $z$, in the geometric painting feature space. The edge between artists is thus considered an expression of similarity between their paintings. These similarities can be studied to investigate the ``influences'' artists exerted over each other. In this way, a form of \textit{historical knowledge discovery} is accomplished, showing (possibly new) influences
among artists.

Graph theory, in fact, provides a powerful framework for investigating the components of such a network and their interactions \cite{deo2017graph}. Its use spans across a number of disciplines including physics, biology, electrical engineering, and so on. Some traditional metrics suitable for describing topological properties of a network are node degree, and closeness and betweenness centrality. 

The degree of a node $v$, denoted as $deg(v)$, is simply the number of edges that are incident to $v$. A node with high degree is a highly connected node. In the specific domain, the degree of each (painter) node $v$ represents the number of painters who are directly linked to $v$. The higher $deg(v)$, the more connected to other painters $v$ is. In other words, degree can represent direct influences.

The closeness centrality of a node $v$ in a graph with $n$ nodes is defined as \cite{freeman1979centrality}:
\[
C(v) = \frac{n-1}{\sum_{u=1}^{n-1}d(u, v)},
\]
where $d(u, v)$ is the shortest path distance between $u$ and $v$. Hence centrality is the reciprocal of the sum of the shortest path distances from $v$ to all other $n-1$ nodes. Since the sum of distances depends on the number of nodes in the graph, this quantity is normalized by the sum of minimum possible distances $n-1$. Closeness centrality indicates whether a node is within a short average distance from every other reachable node in the network, providing information on the ease with which a component of the network can connect to all other components. Thus, the more central a node is, the \textit{closer} it is to all other nodes. Measures of \textit{centrality} are used to identify the most important vertexes in a graph. In the visual art domain, centrality provides an indication of the most influential painter in the artists' network regardless of direct links.

Another measure of centrality is betweenness. For a node $v$, it is the sum of the fraction of all shortest paths passing through $v$, so it is defined as \cite{brandes2008variants}:
\[
B(v) = \sum_{s,t \in V} \frac{\sigma(s,t \mid v)}{\sigma(s,t)},
\]
where $V$ is the set of nodes, $\sigma(s,t)$ is the number of shortest paths between $s$ and $t$, and $\sigma(s,t \mid v)$ is the number of shortest paths through $v$ other than $s$ and $t$. Betweenness centrality of a node $v$ measures the importance of $v$ for the information flow through the network. A large betweenness centrality of a node indicates that many shortest paths between other node pairs pass through that node. Nodes with high betweenness generally connect modules of the network that may become disconnected if these nodes are removed.
In the specific context, a node with high betweenness can represent a ``bridge'', i.e.~a painter who is in the historical influence path between painting schools that could not have been influenced without the existence of that node.

\section{Experimental evaluation}
\label{evaluation}

The effectiveness of the proposed method was investigated on a database collecting paintings of $50$ very popular painters. More precisely, data provided by the Kaggle platform,\footnote{\url{https://www.kaggle.com/ikarus777/best-artworks-of-all-time}} scraped from an art challenge website,\footnote{\url{http://artchallenge.ru}} were used. Artists belong to very different epochs and painting schools, ranging from Giotto di Bondone and Renaissance painters such as Leonardo da Vinci and Michelangelo, to Modern Art exponents, including Pablo Picasso, Salvador Dal\'i, and so on. In particular, nine periods can be recognized: Gothic, Renaissance, Baroque, Romanticism, Impressionism, Post-impressionism, Expressionism, Surrealism, Art Nouveau/Modern Art. Figure \ref{fig:samples} provides sample images of the dataset employed; while, Table \ref{tab:dataset} lists the $50$ painters it includes. Painting images are non-uniformly distributed among painters for a total of $8,446$ images of different sizes. 

Experiments were run on an Intel Core i5 equipped with the NVIDIA GeForce MX110, with dedicated memory of 2GB. As deep learning framework, TensorFlow 2.0 and the Keras API were used. As a tool to perform the graph analysis, Cytoscape was used. It is worth noting that an execution time analysis was not performed. In fact, a key advantage of the proposed method is that its most expensive part, i.e.~the VGG-based feature extraction, can be done completely offline, thus making the visual link retrieval, i.e.~the search over the reduced feature space, dependent only on the collection size.

\begin{table}[!t]
\caption{Painters in the considered dataset.}
\label{tab:dataset}       
\begin{tabular}{ll}
\hline\noalign{\smallskip}
Albrecht Duerer (1471--1528) & Alfred Sisley (1839--1899) \\
\hline\noalign{\smallskip}
Amedeo Modigliani (1884--1920) & Andrey Rublyov (1360--1430) \\
\hline\noalign{\smallskip}
Andy Warhol (1928--1987) & Camille Pissarro (1830--1903) \\
\hline\noalign{\smallskip}
Caravaggio (1571--1610) & Claude Monet (1840--1926) \\
\hline\noalign{\smallskip}
Diego Rivera (1886--1957) & Diego Vel\'azquez (1599--1660) \\
\hline\noalign{\smallskip}
Edgar Degas (1834--1917) & \'Edouard Manet (1832--1883) \\
\hline\noalign{\smallskip}
Edvard Munch (1863--1944) & El Greco (1541--1614) \\
\hline\noalign{\smallskip}
Eugène Delacroix (1798--1863) & Francisco Goya (1746--1828) \\
\hline\noalign{\smallskip}
Frida Kahlo (1907--1954) & Georges Seurat (1859--1891) \\
\hline\noalign{\smallskip}
Giotto di Bondone (1267--1337) & Gustav Klimt (1862--1918) \\
\hline\noalign{\smallskip}
Gustave Courbet (1819--1877) & Henri de Toulouse-Lautrec (1864--1901) \\
\hline\noalign{\smallskip}
Henri Matisse (1869--1954) & Henri Rousseau (1844--1910) \\
\hline\noalign{\smallskip}
Hieronymus Bosch (1453--1516) & Jackson Pollock (1912--1956) \\
\hline\noalign{\smallskip}
Jan van Eyck (1390--1441) & Joan Mir\'o (1893--1983) \\
\hline\noalign{\smallskip}
Kazimir Malevich (1879--1935) & Leonardo Da Vinci (1452--1519) \\
\hline\noalign{\smallskip}
Marc Chagall (1887--1985) & Michelangelo (1475--1564) \\
\hline\noalign{\smallskip}
Mikhail Vrubel (1856--1910) & Pablo Picasso (1881--1973) \\
\hline\noalign{\smallskip}
Paul Cézanne (1839--1906) & Paul Gauguin (1848--1903) \\
\hline\noalign{\smallskip}
Paul Klee (1879--1940) & Pieter Paul Rubens (1577--1640) \\
\hline\noalign{\smallskip}
Pierre-Auguste Renoir (1841--1919) & Piet Mondrian (1872--1944) \\
\hline\noalign{\smallskip}
Pieter Bruegel (1525--1569) & Raffaello (1483--1520) \\
\hline\noalign{\smallskip}
Rembrandt (1606--1669) & René Magritte (1898--1967) \\
\hline\noalign{\smallskip}
Salvador Dal\'i (1904--1989) & Sandro Botticelli (1445--1510) \\
\hline\noalign{\smallskip}
Tiziano (1490--1576) &  Vasilij Kandinskij (1866--1944) \\
\hline\noalign{\smallskip}
Vincent van Gogh (1853--1890) & William Turner (1775--1851) \\
\noalign{\smallskip}\hline
\end{tabular}
\end{table}

\subsection{Visual link retrieval}
\label{eval:visual}

\begin{figure*}[t]
    \centering
    \includegraphics[width=.8\columnwidth]{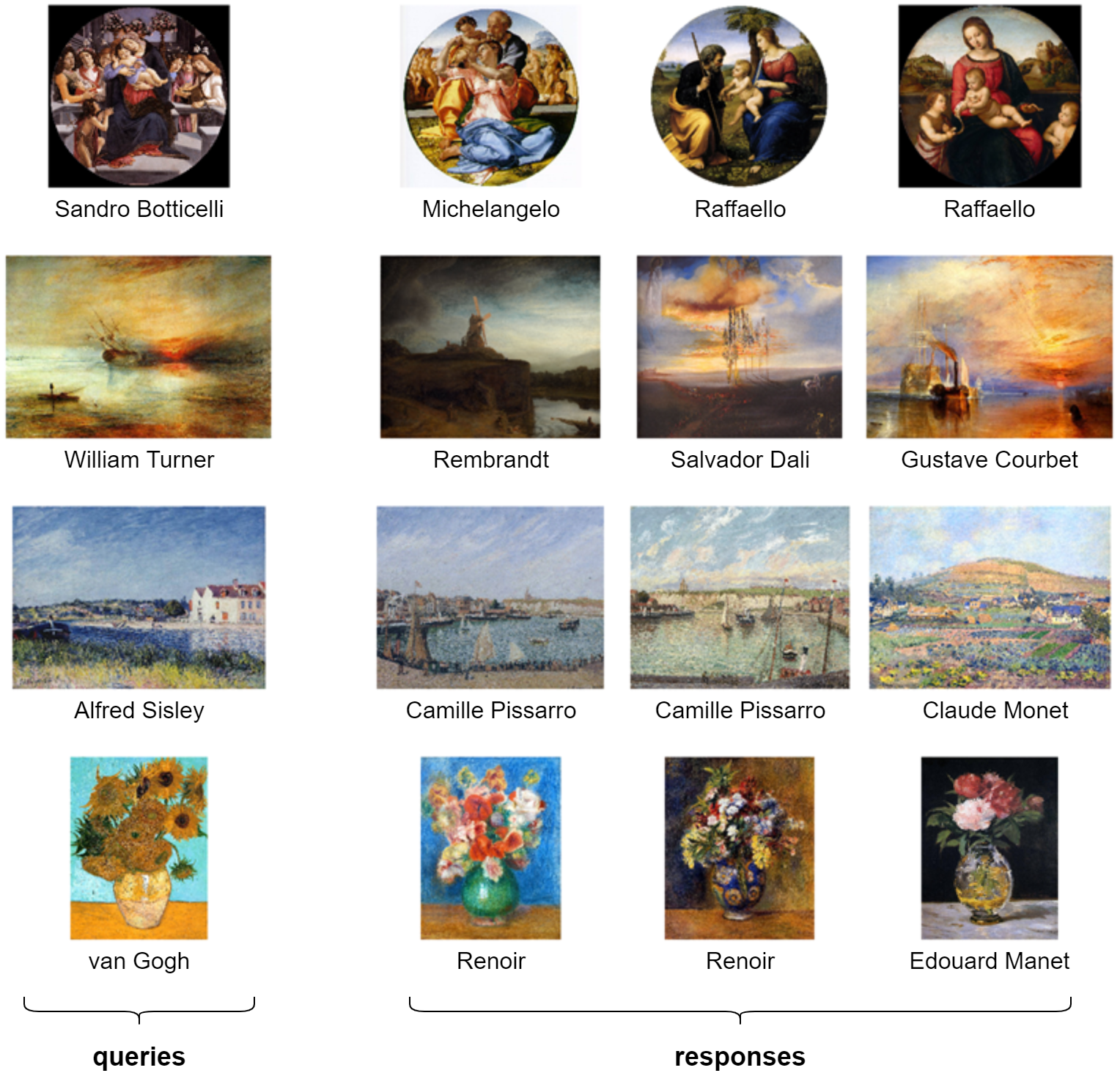}
    \caption{Sample artwork queries and corresponding visually linked paintings provided by the system.}
    \label{fig:responses}
\end{figure*}

Once the reduced features representing paintings were obtained, the Nearest Neighbour matching mechanism was applied to derive, for each query image, the top $k$ matching images ($k=3$ in this case). To give an illustrative example of the system's behavior, Fig. \ref{fig:responses} provides four sample image queries, along with the corresponding top visually linked artworks retrieved by the system. For each query, a brief description of the results is provided below:

\begin{figure}[t]
    \centering
    \includegraphics[width=.75\textwidth]{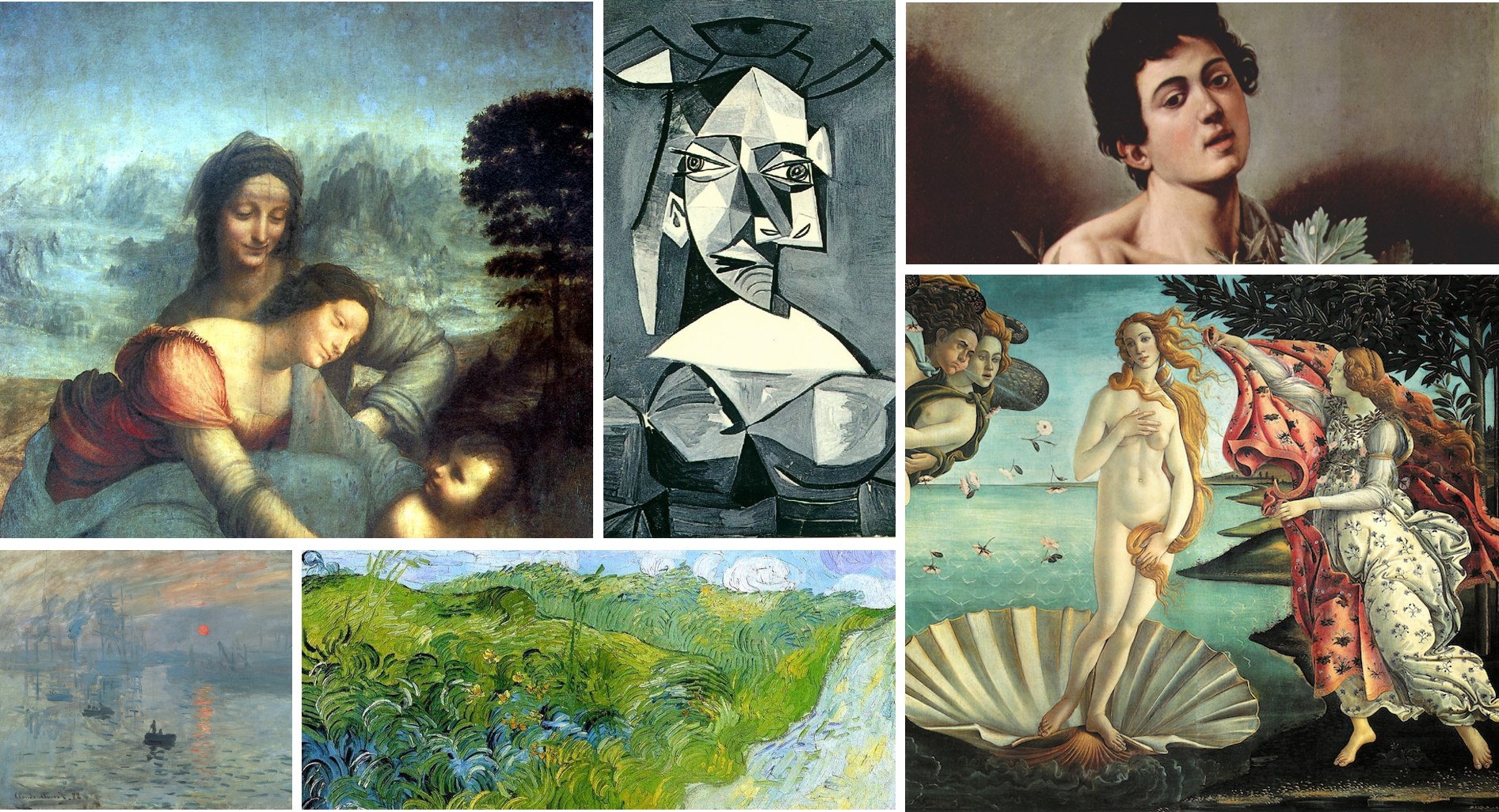}
    \caption{Samples images from the dataset used.}
    \label{fig:samples}
\end{figure}

\begin{enumerate}
    \item[Q1] The first query was the classic ``Virgin and Child with Six Angels and the Baptist'' by the Renaissance artist Sandro Botticelli. As expected, the visual features provided by the deep network were useful in retrieving paintings similar in composition (holy family) and shape (tondo); 
    \item[Q2]
    The second image query was the Romanticist ``Fort Vimieux'' by William Turner, depicting a classic red sunset of the author. It can be seen that the system was able to retrieve paintings similar in both content and color distribution; 
    
    \item[Q3]
    The third query was the Impressionist ``Confluence of the Seine and the Loing'' by Alfred Sisley. It can be noted that the three neighbors, i.e.~two artworks by Camille Pissarro and a work by Claude Monet, share the same painting style, characterized by the typical color vibration;
    
    \item[Q4]
    Finally, a version of the ``Sunflowers'' series by Vincent van Gogh was considered as a query. As expected, the top three images retrieved by the system represent still lifes, two by Renoir, the other by \'Edouard Manet.
\end{enumerate}

Overall, based on a qualitative evaluation of the retrieval results, it can be concluded that the proposed system is capable of finding visual links that are not in contrast with the human perception. The visual links discovered by the system are sufficiently justifiable by a human observer and in most cases resemble the intrinsic criteria humans adopt to link visual arts. These criteria combine visual elements, such as colors and shapes, and conceptual elements, such as subject matter and meaning of the painted scene.

As for a quantitative evaluation, it is worth noting that the proposed method is completely unsupervised, so there is no ground truth about the visual links available. For this reason, it is difficult to obtain a quantitative assessment of the effectiveness of the proposed method. In particular, evaluating the recall of the method is unfeasible, due to the difficulty of establishing the set of all meaningful links. On the contrary, an evaluation of its precision is practical. To this end, five art experts were involved to obtain a subjective evaluation of the links provided by the system. $100$ images were randomly selected as queries, thus creating a pool of $100$ pairs of the form $(\texttt{q}_i,\texttt{p}_i)$, where $\texttt{q}_i$ was the $i$th query and $\texttt{p}_i$ was a painting chosen at random from the top three matching paintings retrieved by the system for query $\texttt{q}_i$. Each expert was asked to examine each pair $(\texttt{q},\texttt{p})$ and establish whether the semantic link between the query $\texttt{q}$ and the artwork $\texttt{p}$ was meaningful or not. 
It should be noted that experts were not constrained to adopt a specific meaning of ``link''. They were only asked to tell if the retrieved visual links made sense according to their experience/background/perception. To avoid the effects of influence among experts, they had to perform the task blindly with each other. An uneven number of experts was involved so that, for each single link, the mode of the evaluations provided by the experts was considered. This test yielded $72$ meaningful visual links out of the $100$ randomly proposed links, resulting in a precision of $72\%$. Although this value appears to be not as high as the precision values usually reported in the literature on information retrieval, it is quite encouraging considering that, given a query, the system is forced to retrieve similar images even when similar artworks are actually missing in the painting collection. 
%

Overall, the results obtained show that the proposed method was able to find a suitable model so that, once visual features are automatically extracted from painting images, it can help to acquire new knowledge about relationships among paintings, useful for several applications. In particular, the proposed method benefits from the capability of Convolutional Neural Network models to exploit complex nonlinear relationships within data.

\subsection{Knowledge discovery}
\label{eval:discovery}

\begin{figure*}[t]
    \centering
    \includegraphics[width=\columnwidth]{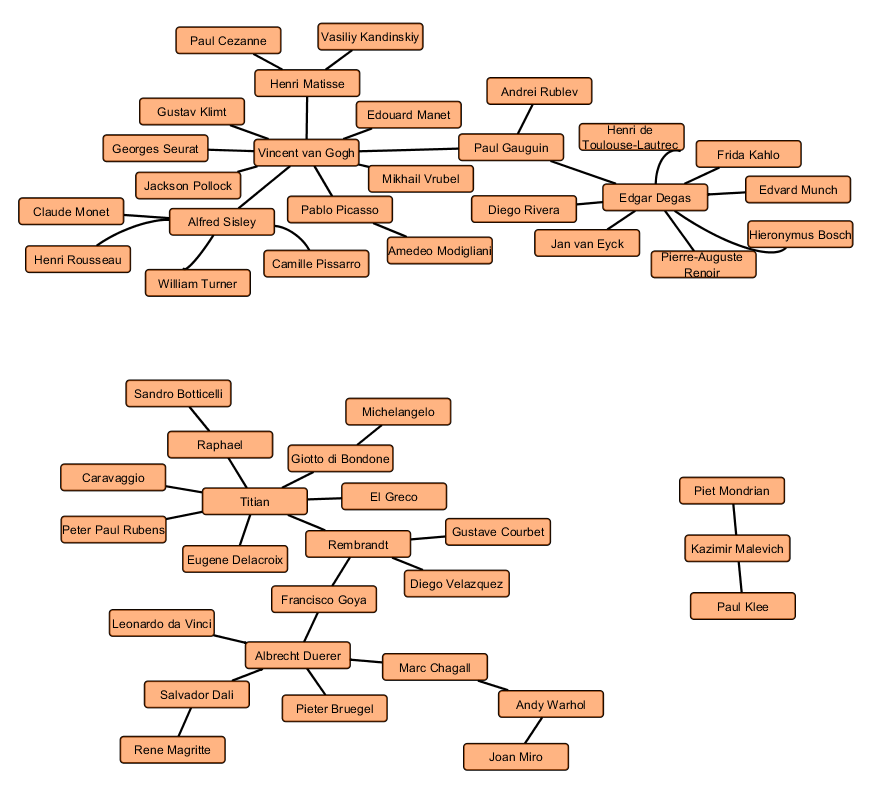}
    \caption{Influence graph between painters. The connections between painter nodes are established on the basis of the visual similarity between their artworks. Homogeneous groups that share some stylistic characteristics are clearly recognizable.}
    \label{fig:network}
\end{figure*}

Figure \ref{fig:network} represents the graph of painters built on the retrieved links. It is easy to see that the graph is made up of three connected components isolated one from each other. The largest connected component is characterized by Modern Art exponents, including Impressionist and Post-Impressionist painters, such as Edgar Degas, Paul Gauguin and Vincent van Gogh. The second largest sub-graph, on the other hand, is characterized by more classic painters, mostly belonging to the Renaissance period, such as Tiziano and Albrecht Duerer. Finally, the smallest connected component includes three painters who pioneered the Abstract Arts of the first half of the $20$th century, namely Piet Mondrian, Kazimir Malevich and Paul Klee. In other words, the network analysis was able to reveal, within the painting collection, three clusters, each characterized by homogeneous features.

A more refined analysis can be done at the node level. By looking at Table \ref{tab:artists}, which shows the most important nodes in descending order of node degree, it can be seen that the most ``influential'' artists correspond to the \textit{hubs} of the network. Not surprisingly, the most important node appears to be Vincent van Gogh, with a degree of $9$, a closeness centrality of $0.52$ and a betweenness centrality of $0.80$. It can be seen that van Gogh has only two degrees of separation with several artists, mostly belonging to the Impressionist period. Moreover, through Paul Gauguin, who has the top second closeness centrality, van Gogh is connected to another cluster composed mainly by Post-Impressionist and Expressionist painters. This was expected as van Gogh was one of the most famous and influential figures in all of Western Art. He was one of the most prolific painters with around $860$ oil paintings, including still lifes, landscapes, portraits and self-portraits. Unfortunately, he was not successfully in life, committing suicide at a young age after years of poverty. 

The second most influential artist in accordance with the proposed analysis turns out to be Tiziano, who shows the top second betweenness centrality, that is $0.64$, among the other nodes. He was an Italian painter and one of the most prominent members of the so-called Venetian School during the $16$th century. Tiziano focused mainly on mythological and religious subjects. It is generally believed that his painting style had a strong influence not only on other painters of the Italian Renaissance, but also on future generations of Western Art.

Another important node appears to be Edgar Degas. This result was also not surprising, as Degas is generally considered one of the founders of Impressionism (although, during his life, he preferred to be considered a Realist). He particularly mastered the depiction of movement, as can be seen in several of his works which portray dancers.

It is worth noting that incredibly famous artists such as Leonardo Da Vinci and Michelangelo have very low degree (i.e., $1$) and they are at the boundaries of their respective sub-graphs, although they are unanimously considered to be two universal genius. Their interests and curiosity, in fact, spanned across a wide range of disciplines, including not only art, but also literature, architecture and science. The result obtained can be explained considering that, although their works certainly influenced the Western culture in general, Leonardo Da Vinci and Michelangelo were not as prolific in painting as other popular artists. Indeed, as far as arts are concerned, they mainly focused on drawings and sculptures, respectively, rather than paintings.

As a further remark, it can be observed in Fig.~\ref{fig:responses} that a landscape of Rembrandt was found similar to a work by Salvador Dal\'i, even though the two artists are not directly linked in the graph. This is because the painters' graph is obtained by linking painters based on the frequency of the top one visual links retrieved among their paintings. In other words, a link in the graph reflects a \textit{global} similarity between their artistic production, without depending on very specific visual links.

\begin{table}[t]
\caption{Graph-based analysis of the top painters.}
\label{tab:artists}       
\begin{tabular}{lccc}
\hline\noalign{\smallskip}
Painter & Degree & Closeness & Betweenness \\
\noalign{\smallskip}\hline\noalign{\smallskip}
Vincent van Gogh & $9$ & $0.52$ & $0.80$ \\
Edgar Degas & $8$ & $0.39$ & $0.49$ \\
Tiziano & $7$ & $0.38$ & $0.64$ \\
Albrecht Duerer & $5$ & $0.35$ & $0.56$ \\
Alfred Sisley & $5$ & $0.39$ & $0.30$ \\
Rembrandt & $4$ & $0.40$ & $0.62$ \\
Henri Matisse & $3$ & $0.36$ & $0.15$ \\
Paul Gauguin & $3$ & $0.46$ & $0.50$ \\
\noalign{\smallskip}\hline
\end{tabular}
\end{table}

Finally, the artistic influences found by the proposed method have been qualitatively compared with the findings reported in the work by Saleh et al.~\cite{saleh2016toward}, which is based on traditional features extracted from paintings of very popular artists. It can be seen that many of the influences suggested by their method are consistent with the results reported in this paper. For example, strong links were suggested between Delacroix, Rubens, Tiziano, Raffaello and El Greco: in fact, in the proposed graph, these artists form a recognizable cluster with direct links between them. Another relationship was found by the authors between Leonardo and Duerer: this is also a finding of the present work. Indeed, it is known that the two artists were in contact at the time. Another pattern that is in common between this work and the results of Saleh et al.~is the strong connection found between Monet, Sisley and Pissarro. Similarly, a clear path emerges in the graph linking Goya, Rembrandt and Velazquez: influences between these painters were also suggested in the work cited. Analogous considerations also apply to the relationships found between Kahlo and Renoir, and Manet, Picasso and Degas. Interestingly, also in \cite{saleh2016toward} the authors found a relationship between Malevich and Mondrian. In the graph presented, they form a small connected component together with a third artist, namely Paul Klee. However, in \cite{saleh2016toward} Klee does not appear to be directly connected to the other two artists, as found in this work.

\section{Discussion}

The proposed method can be advantageous not only for art historians but also for other figures. For example, enthusiasts can benefit from the automatic link retrieval when visiting digital collections of museums and art galleries online. This can favor a sort of interactive navigation able to promote the fruition of art. The same logic can be applied to physical museums: curators, in fact, could use applications of the proposed tool to enrich the visiting experience. For example, once the particular interest of a visitor to an artwork is confirmed, the system may recommend similar works the visitor may be interested in. This strategy can also be used in conjunction with Internet-of-Things sensors \cite{bharti2020optimized,bharti2020optimal} to improve the visiting experience of cultural sites. The proposed method can also be useful to assist art experts in detecting plagiarism. In fact, when the similarity among two paintings exceeds a given threshold, the method can indicate a suspected plagiarism. Finally, we observe that the proposed knowledge discovery methodology may be extended to domains other than the artistic one, if a recognizable semantics can be attributed to the visual links between images.

A limitation of the present study is the small size of the dataset adopted, which hampers the  possibility of generalizing the results obtained. The choice of the dataset was made in the expectation that easy to interpret results could have been obtained by exploring the visual links among very famous painters. In fact, despite these constraints, the reported evaluation is very promising and the results of this study are expected to make way for a working system in the cultural heritage settings. Extending the proposed evaluation to other, larger datasets is currently the topic of future research.

\section{Conclusion}
\label{conclusion}

Determining visual similarities among paintings, as well as influences among artists, is an intrinsically subjective task for human experts and depends on several factors, most notably their aesthetic perception. To help experts with an automatic method, this paper has tackled the problem of making a machine capable of mimicking this complex perception. 
Since the proposed method only works in an unsupervised fashion, one of its key advantages is that it relies solely on the visual attributes extracted by a deep CNN, without the need for additional metadata, which are typically very difficult to collect. For the same reason, another key aspect of the proposed approach is its efficiency, since the most expensive stage, i.e.~the CNN-based feature extraction, can be done completely offline. 
The results obtained are encouraging for the purposes of the present research, whose long-term goal concerns the automatic discovery of patterns in painting images without the need of prior knowledge. 


%
\section*{Conflict of interest}

The authors declare they have no conflict of interest.

\section*{Acknowledgement}

Gennaro Vessio acknowledges funding support from the Italian Ministry of Education, University and Research through the PON AIM 1852414 project.

\bibliographystyle{ieeetr}
\bibliography{biblio.bib}   

%
%

\end{document}